\documentclass[journal]{IEEEtran}

\usepackage{cite}

\ifCLASSINFOpdf
\else
\fi

\usepackage{amsmath}
\usepackage{multirow}
\usepackage{subfig}
\usepackage{graphicx}
\usepackage{adjustbox}
\usepackage{xcolor}
\usepackage{amsfonts}

\begin{document}

\title{Physics Augmented Tuple Transformer for Autism Severity Level Detection}



\author{Chinthaka Ranasingha,
        Harshala Gammulle, ~\IEEEmembership{Member,~IEEE,}
        Tharindu~Fernando,~\IEEEmembership{Member,~IEEE,}
        Sridha~Sridharan,~\IEEEmembership{Life Senior Member,~IEEE,}
        ~and~ Clinton~Fookes,~\IEEEmembership{Senior Member,~IEEE.}
     
\IEEEcompsocitemizethanks{\IEEEcompsocthanksitem C. Ranasingha, H. Gammulle, T. Fernando, S. Sridharan, and C. Fookes are with The Signal Processing, Artificial Intelligence and Vision Technologies (SAIVT), Queensland University of Technology, Australia.\protect\\ }}

\markboth{Journal of \LaTeX\ Class Files,~Vol.~14, No.~8, April~2024}%
{Shell \MakeLowercase{\textit{et al.}}: Bare Demo of IEEEtran.cls for IEEE Transactions on Magnetics Journals}

\IEEEtitleabstractindextext{%
\begin{abstract}
Early diagnosis of Autism Spectrum Disorder (ASD) is an effective and favorable step towards enhancing the health and well-being of children with ASD. Manual ASD diagnosis testing is labor-intensive, complex, and prone to human error due to several factors contaminating the results. This paper proposes a novel framework that exploits the laws of physics for ASD severity recognition. The proposed physics-informed neural network architecture encodes the behaviour of the subject extracted by observing a part of the skeleton-based motion trajectory in a higher dimensional latent space. Two decoders, namely physics-based and non-physics-based decoder, use this latent embedding and predict the future motion patterns. The physics branch leverages the laws of physics that apply to a skeleton sequence in the prediction process while the non-physics-based branch is optimised to minimise the difference between the predicted and actual motion of the subject. A classifier also leverages the same latent space embeddings to recognise the ASD severity. This dual generative objective explicitly forces the network to compare the actual behaviour of the subject with the general normal behaviour of children that are governed by the laws of physics, aiding the ASD recognition task.  The proposed method attains state-of-the-art performance on multiple ASD diagnosis benchmarks. To illustrate the utility of the proposed framework beyond the task ASD diagnosis, we conduct a third experiment using a publicly available benchmark for the task of fall prediction and demonstrate the superiority of our model.
\end{abstract}

\begin{IEEEkeywords}
Autism Severity Level Detection, Physics Informed Neural Networks, Transformers, Deep Neural Networks. 
\end{IEEEkeywords}}

\maketitle

\IEEEdisplaynontitleabstractindextext

\IEEEpeerreviewmaketitle

\section{Introduction}
\label{sec:introduction}
\IEEEPARstart{A}{utism} Spectrum Disorder (ASD), also known as autism, is a neurodevelopmental condition that poses underlying communication and behavioral challenges and commonly co-occurs with cognitive conditions of patients \cite{whatisautism, wadhera2022brain}. Because of the significant variations of the symptoms and types between individuals, ASD is referred to as a spectrum disease. Learning, thinking, and problem-solving skills in people with ASD can range from exceptionally gifted to severely challenged. They may also behave, speak, and learn differently from most other people \cite{employment_for_autism}. As such, early diagnosis of autism symptoms is vital and the Autism Diagnostic Observation Schedule (ADOS) is one of the widely used metrics to measure the severity of autism \cite{ados}. 

Early diagnosis of ASD is a feasible and useful step toward enhancing the health of those with autism and can help to reduce the stress on the families who are supporting them \cite{early_identification, ji2023automated}. Some autism diagnostic studies have examined the possible advantages of addressing the social and communication aspect with the help of therapeutic interventions while some others focused on certain medications for core symptoms \cite{behavioral_psychosocial,GaoWearable2024,Lalawat2024Anautomatic}. One of the most widely researched and clinically supported therapeutic interventions for individuals with ASD is Applied behavior analysis (ABA) which aims to improve the behaviour of the by using systematic approaches \cite{aba}. 

A specific ABA approach named discrete trial training (DTT) has been widely used in autism  interventions in which a child is given a specific instruction or discriminative stimulus by a therapist \cite{dtt}. If the child performs the expected behavior, they receive a reward. However, if the child does not respond appropriately, the therapist could use cues or demonstrations to change the behavior. This structured approach helps reinforce desired behaviors and provides guidance when necessary. In recent years, robots have been used to assist the child during DTT. 
During this alternative form of therapy known as Robot Enhanced Therapy (RET) \cite{ret1}, the robot actively assists the child through a game-like activity, and at the same time the therapist observes and guides the child by observing the interaction of the child with the robot. Information gathered during these sessions could be used to assess the ADOS level. However, manual analysis of ADOS is a challenging task that requires specialised knowledge and is prone to human error. As a solution, a number of machine learning-based methods \cite{CaiSensing2019, attention-based-autism-screening-modality, zahan2023human, 3DHumanSensing, VIDEO-BASEDEARLYASDDETECTIONVIATEMPORALPYRAMIDNETWORKS, wei2023vision, VideoGestureAnalysisForASDDetection} have been proposed to assess ADOS levels automatically.

Despite the existence of numerous architectures for automated ADOS-level classification, these methods do not achieve the required levels of robustness or efficiency for practical application. Firstly, facial expression \cite{facialimage, zitouni2022lstm}, and eye gaze \cite{eyegaze} features have been utilised in most of the prior works which make their applicability highly constrained due to privacy concerns and challenges with respect to collecting those data. Secondly, the limited number of skeleton-based methods \cite{zahan2023human}, have complex architectures and require substantial computational power to execute these algorithms, which is not available in clinical settings. Moreover, existing algorithms \cite{zahan2023human} do not generalise well to diverse ADOS levels.

\begin{figure*}
    \centering
    \includegraphics[width=1\linewidth]{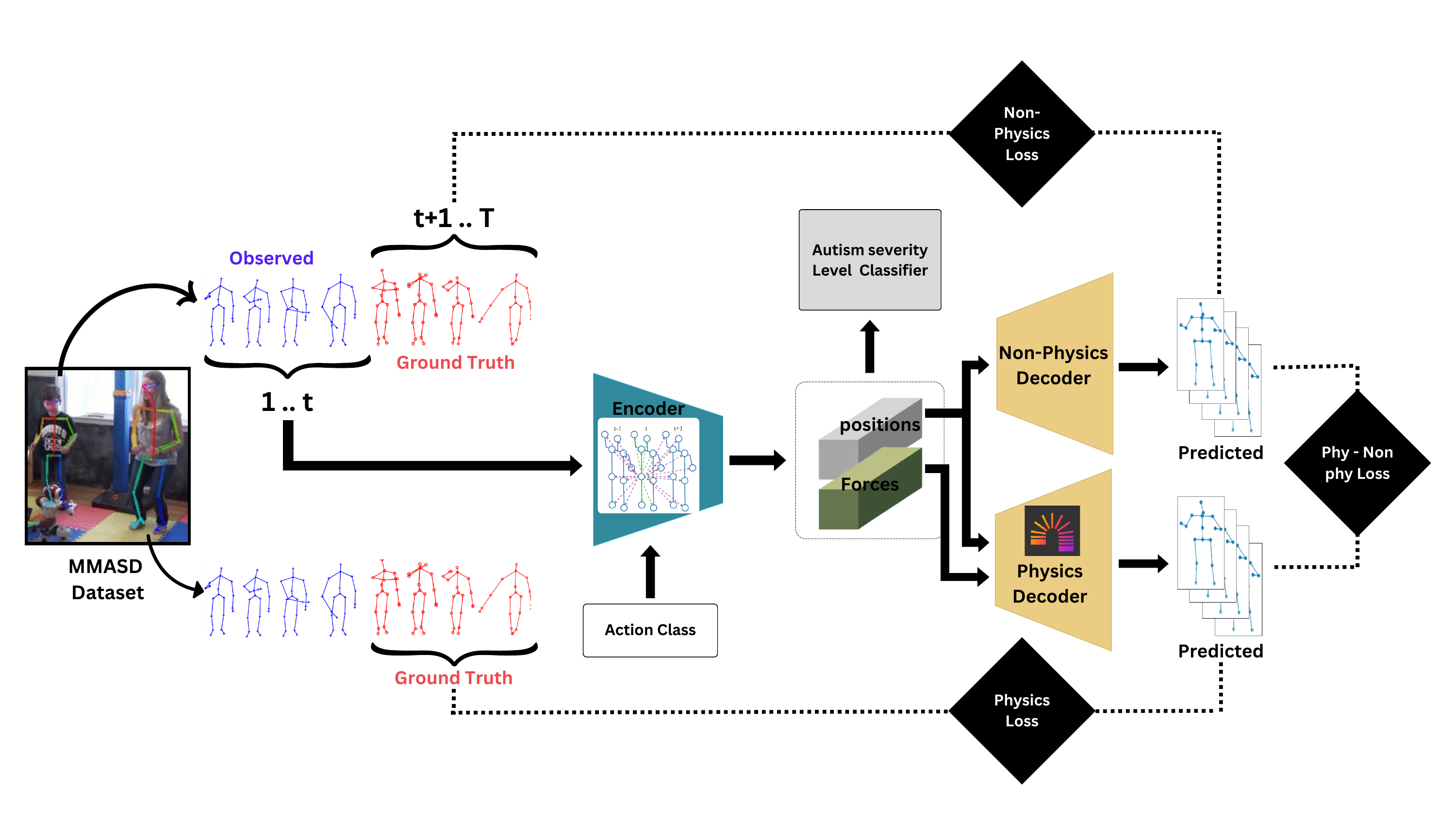}
    \caption{\textbf{Illustration of Physics Augmented Tuple Transformer (PATT) framework.} It contains a transformer encoder, a classifier for Autism Severity Level Detection and two decoders namely non-physics decoder and physics decoder to generate future physical representations. Input to our framework is the type of the action that the subject is performing and the observed skeleton sequence (1..t). Encoder encodes this information as latent  position and force embedding. Non-physics decoder generate future skeleton sequence (t+1..T) by leveraging only the encoded positions. Physics decoder also generates future skeleton sequence (t+1..T) by leveraging both encoded positions and forces. When providing ground truths to as a supervision signal the non-physics decoder receives subject's skeleton sequence which contains different ASD levels. However the physics decoder receives the TD sequence of the same action performed by the therapist. As such our network gains the ability to compare the anomalous ASD and TD behaviour and identify the autism levels. Our physics decoder helps this normal-abnormal comparisons by decoding the future normal behavior of the TD subject utilising the laws of physics.} 
    \label{fig:simplified_architecture}
\end{figure*}

To address these limitations this paper proposes a novel physics-informed neural network architecture for skeleton-based ASD severity recognition by predicting ADOS scores.

To be more precise, our architecture analyses the laws of physics that relate to the skeleton sequences in conjunction with the tasks that the subject is expected to do, to distinguish between various ADOS levels.
The resultant architecture (See Figure \ref{fig:simplified_architecture}) is lightweight, interpretable, robust, and generalisable across multiple datasets. The main contributions of our paper can be summarised as follows:

\begin{enumerate}
    \item We introduce a novel lightweight architecture that learns ADOS scores by discriminating between highly probable future poses based on the laws of physics and the actual poses of the subject.

    \item We demonstrate how the laws of physics could be used to differentiate between normal and abnormal future behaviour patterns, allowing effective localisation of behavioral traits that are indicative of ASD.

    \item We present a new discriminative loss that further improves learning by differentiating between TD and ASD behaviors.

    \item We evaluate the proposed approach using MMASD \cite{mmasd} and DREAM \cite{dream} benchmark datasets and demonstrate state-of-the-art performance. To demonstrate the utility of the proposed framework beyond the detection of ASD, we show how it can be deployed for the fall prediction task using the URFD \cite{urfd} dataset and achieve state-of-the-art performance.
    
\end{enumerate}

In Sec.\ref{sec:related_work} we review related work.  In Sec. \ref{sec:methodology} we describe the proposed architecture. In Sec. \ref{sec:eval}, we provide the details of the datasets, and evaluation metrics used in this research along with the implementation details and results. Sec. \ref{sec:ablation} contains ablation studies to assess the relative contribution of the components of the proposed framework and in Sec. \ref{sec:conclusion} we summarize the main conclusions.

\section{Related Work} \label{sec:related_work}

\subsection{Automated ADOS Prediction Approaches}

When considering the literature on automated ADOS level identification, eye-gaze patterns, and gestures have been the most commonly used factors for the analysis. 

Using gaze behaviours it has been experimentally validated that there is a decrease of attention in children with ASD compared to TD children \cite{Lookingatmoviesandcartoons}. Motivated by this observation, Li et al. \cite{Appearancebasedgazeestimation} proposed a computational framework that leverages gaze from a raw video to estimate autism levels. Liu et al. \cite{autism_research_2016} used face-scanning patterns to identify children with ASD by adopting classification machine learning techniques. Similarly, Wang et al. \cite{Atypicalvisualsailancey} proposed to train a model to classify individuals with ASD based on the eye gaze patterns during image-viewing. Jiang and Zhao \cite{Learningvisualattention} extended the work in \cite{Atypicalvisualsailancey} by adopting eye-tracing and deep neural networks for ASD screening. However, in a clinical setting it is difficult to obtain eye gaze patterns as eye tracking requires specialised hardware.

In another line of work, gesture patterns have been analysed to discriminate between children with ASD and TD children. For instance, Anzulewicz et al.\cite{motor} has computationally assessed autism based on the recorded motor patterns during smart device gameplay and concluded that there is a wide difference between the hand gesture patterns of ASD and TD children. Another work that utilises motor function to discriminate between ASD and TD is the recurrent deep neural network-based approach proposed by Zunino et al. \cite{VideoGestureAnalysisForASDDetection} in which they have analysed the grasping action in videos. 

Our work is inspired by the recent success of spatio-temporal behaviour analysis models that leverage the full body motion of the subject to identify motion cues that are indicative of ASD. Negin et al. \cite{Vision-assistedrecognition} used the Bag-of-Visual-Words approach to develop several action recognition-based frameworks to identify ASD characteristics in children. Tian et al. \cite{VIDEO-BASEDEARLYASDDETECTIONVIATEMPORALPYRAMIDNETWORKS} proposed an end-to-end deep architecture for video-based early ASD detection. They have utilized a temporal pyramid network to capture high-level semantics temporal feature maps at all scales and a discriminator to detect high-risk repetitive behaviours. Similarly in \cite{wei2023vision}, state-of-the-art human action recognition architectures have been adapted for the task of recognising ASD. Marinoiu et al. \cite{3DHumanSensing} have used 3d human pose data for action and emotion recognition in children with ASD. Specifically, 2D and 3D pose features and interactions between the child and therapist have been leveraged as motion features. A hierarchical bi-directional recurrent neural network has been used to model the pose-related features. 

More recently, \cite{pandey2019guided} proposed a guided weak supervision architecture to address the lack of large-scale annotated datasets when performing automated video-based detection of ASD using supervised machine learning. The authors propose to match the similarities between the actions in publicly available large-scale video action datasets and the actions performed in datasets with children that have ASD symptoms. A multimodal approach for screening ASD subjects is proposed in \cite{attention-based-autism-screening-modality} in which the authors have leveraged both behavioural patterns extracted from video and eye-gaze patterns. Two ResNet feature extractors are employed for extracting features from individual modalities and the temporal relationships across the extracted features are learned through modality-specific LSTM networks. In \cite{zahan2023human}, the authors propose to use gesture and gait information for identifying autism-related behaviour. A Graph Convolution Neural Network (GCN) is used to embed the skeleton information in the latent space. To further augment the learning capabilities the authors propose to encode the skeleton as a skeleton picture element such that a vision transformer can be used to learn the relationships across skeleton joints. 

In contrast to the existing studies on behaviour-based identification of ASD, our work leverages the laws of physics that could be applied to skeleton motion to identify anomalous behaviours. Specifically, we use a state-of-the-art transformer-based encoder to encode the spatio-temporal characteristics of the subject's skeleton in a latent space. The future motion of the subject is decoded by two decoders. One decoder simply uses prediction error-driven optimisation to learn the subject's behaviour while the other decoder uses physics-based laws to predict the future skeleton. A specialised discriminative loss is employed to help emphasise the discriminative patterns between the two decoded outputs. To the best of our knowledge, this is the first work to utilise physics-augmented generative learning for the automated classification of ASD.

\subsection{Transformer-based Skeleton Data Analysis}

Transformers have revolutionised the field of deep learning and are being used as the backbone of many state-of-the-art applications ranging from image classification and recognition \cite{Dosovitskiy2020animage, zha02021transformer}, action recognition \cite{Plizzari2021skeleton} to object detection \cite{beal2020towards}. The attention mechanism \cite{attention} leverages to extract salient information from sequences has enabled it to better capture context and handle long-range dependencies.  

The success of transformers has seeped into the domain of action recognition as well. Video Action Transformer \cite{girdhar2019video} and Recurrent Vision Transformer \cite{yang2022recurring} are some notable works in the video-based action recognition domain. In the biomedical behaviour analysis literature, skeleton data is used widely due to the numerous unique advantages over videos such as compactness and privacy preservation. In the past few years, the GCN \cite{liu2020Disentangling, STGCN} has been widely used as the backbone for extracting useful features from the input skeleton data. The skeleton data itself forms a graph structure, as such the graph neural networks are a natural choice for processing skeleton data. Furthermore, GCNs are highly effective in learning relationships across adjacent nodes enabling the learning of relationships among adjacent joints. 
Processing skeletal data for behavior modeling with transformers is a relatively new adaptation.
Among the limited number of works on transformer-based skeleton data modelling, we would like to compare \cite{ibh2023tempose}, \cite{wang20233mformer}, and \cite{sttformer} together with the proposed transformer architecture. Ibh et al. \cite{ibh2023tempose} proposed a transformer-based framework for fine-grained modelling of interactions among multiple players in team sports. In \cite{wang20233mformer}, Higher-order Transformer embeddings are used to model higher-order dynamics in the skeleton for action recognition. Moreover, in \cite{sttformer} Spatio-Temporal Tuples Transformer (STTFormer) model is proposed which proposes a spatiotemporal tuples self-attention module for efficient modelling of skeleton sequences. Specifically, the proposed attention scheme is capable of capturing the relationship of all joints across multiple consecutive frames at the same time, making it an efficient backbone for modelling skeleton data. Inspired by this we modify the tuple-based encoder architecture in the proposed framework. However, in contrast to \cite{wang20233mformer, ibh2023tempose, sttformer} we incorporate two decoder branches in the proposed method, one of which is driven by the physical laws to aid the recognition task. We further enhance the ability to learn informative features with the aid of proposed discriminative loss.

\section{Methodology}\label{sec:methodology}

This section first describes the overall framework of our proposed physics-augmented tuple transformer (PATT). Our proposed model consists of an encoder followed by two decoders. We introduce the architecture of our encoder in Sec. \ref{sec:encoder}, and the physics and non-physics-based decoders in Sec. \ref{sec:physics_decoder} and \ref{sec:non_physics_decoder}, respectively. We discuss the training loss functions in Sec. \ref{sec:loss}.

\begin{figure*}
    \centering
    \includegraphics[width=1\linewidth]{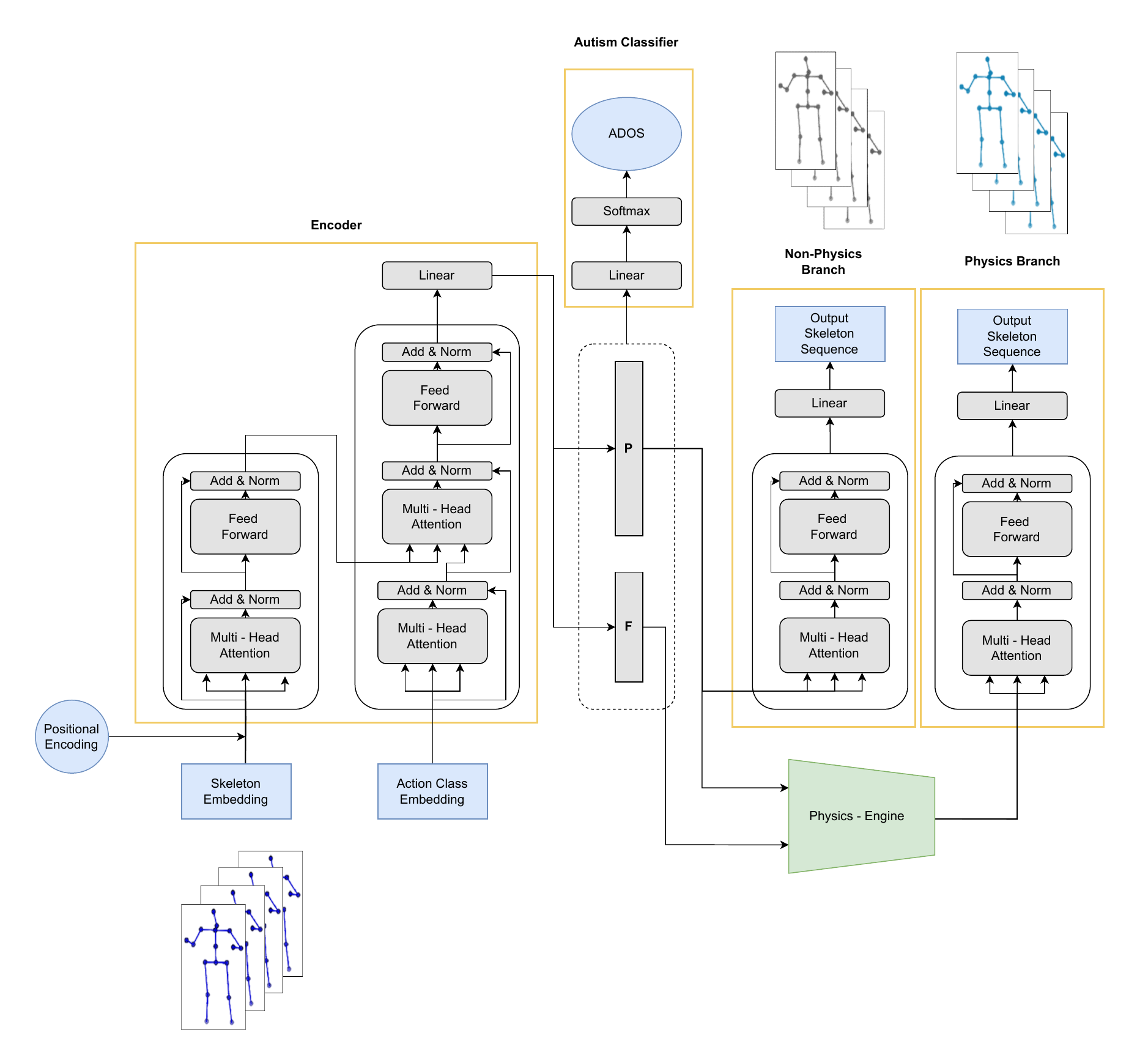}
    \caption{\textbf{Overall framework of Physics-Augmented Tuple Transformer (PATT).} The inputs to the model are the skeleton sequences and the relevant action classes. Initially, the STTFormer-based \cite{sttformer} encoder takes the input and predicts the joint positions (P) and forces (F). Then, the physics-based decoder takes both the generalized positions and forces to predict the next state of the skeleton sequence based on the Lagrangian dynamics, while the non-physics-based decoder takes only generalized positions to generate the next state of the skeleton sequence. The decoders are discarded during the inference and using both generalized positions and forces a simple feed-forward neural network performs the ADOS score prediction. }
    \label{fig:Overall_architecture}
\end{figure*}

\subsection{Overall framework}

The overall framework of the proposed Physics-Augmented Tuple Transformer (PATT) model is shown in \textbf{Figure \ref{fig:Overall_architecture}}. Our architecture follows the encoder-decoder transformer architecture. The inputs for the model are a skeleton sequence, which is composed of human joints, and the corresponding action class.

Inspired by the sequence division step of the STTFormer \cite{sttformer} we adapt the encoder of STTFormer for encoding human skeleton information. This encoding step is computationally efficient and effective in capturing dependencies between different joints among different frames. However instead of directly using the encoding steps of STTFormer, we have simplified the encoding process by further reducing the complexity, which makes the encoding efficient even with a higher number of frames and reduces the overfitting of the model in situations where the data is scarce. Specifically, the final encoding layer of the STTFormer \cite{sttformer} encoding module has been removed and the expansion of the input channels was handled by the input mapping layer.

Our encoder generates the normalized coordinates, the joint positions, and their corresponding forces leveraging the skeleton sequence and the action class as input. This encoded information is utilised by two decoder streams, namely the physics-based decoder and the non-physics-based decoder. The non-physics decoder generates a skeleton sequence of future frames, given only the generalized joint positions. On the other hand, our physics-based decoder predicts the future sequence by using both joint positions and forces. This input difference enforces the separation of the latent embedding into positions and forces where the non-physics branch is purely driven by the coordinates while the physics branch utilises complementary information regarding forces. The predictions generated by physics-based decoder is purely driven by physics-based theories regarding the skeleton structure and the forces. During the inference, both decoders are discarded and a Feed-Forward Neural Network performs autism severity level prediction using the positions and forces as the input. The following subsections illustrate the details of our encoder, decoders, and autism-severity-level classifier in detail.

\begin{figure}
    \centering
    \includegraphics[width=.8\linewidth]{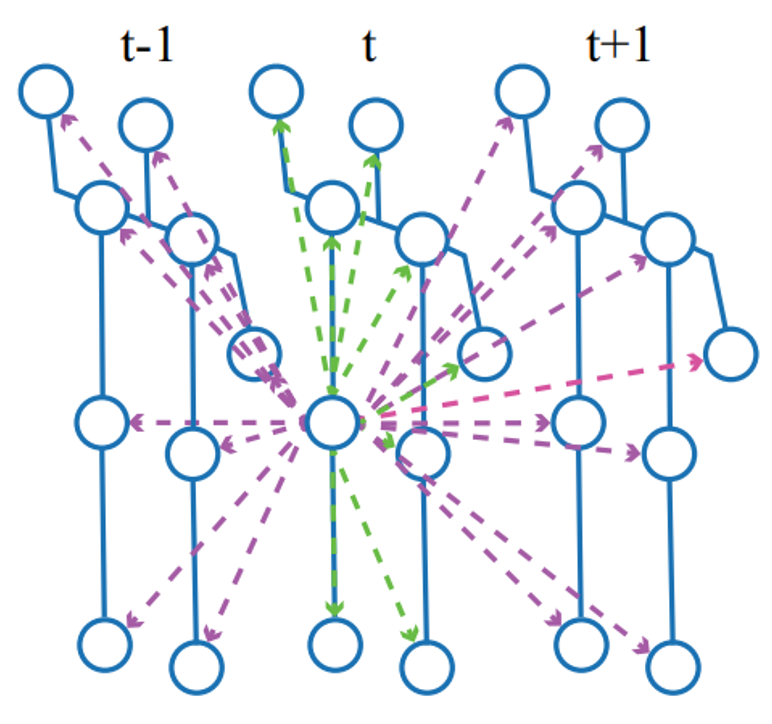}
    \caption{Self-attention scheme that simultaneously captures the relationship between every joint in multiple successive frames.}
    \label{fig:jont_dependancey}
\end{figure}

\subsection{Encoding the Skeleton Sequence and Action Information}\label{sec:encoder}

A crucial aspect of skeleton-based analysis tasks is capturing the inter-dependencies among joints and sequential-dependencies between frames. In addition, we need our encoder to learn the conditional relationship between the encoded sequential dependencies among frames and the action that the subject is supposed to be performing. Transformers have a great potential for modeling these dependencies. Other than capturing the aforementioned dependencies our encoder has the ability to capture the relationship between every joint in successive frames \textbf{(See Figure \ref{fig:jont_dependancey})} by splitting the skeleton sequence into several segments, each containing multiple consecutive frames.

\begin{figure}
    \centering
    \includegraphics[width=.8\linewidth]{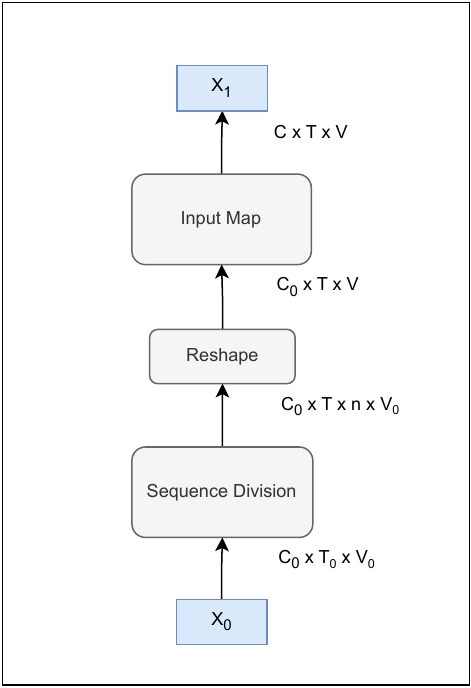}
    \caption{Spatial-temporal tuples encoding module.}
    \label{fig:input_encoding}
\end{figure}

\noindent\textbf{Skeleton Sequence Encoding:} The encoding process is illustrated in Figure \ref{fig:input_encoding}. Initially, the raw skeleton sequence, \(X_0 \in \mathbb{R}^ {C_0\times T_0\times V_0}\), where $C_0$ denotes the dimension of node features , $T_0$ denotes is the length of the input sequence, of $V_0$ joints,  is divided into \(T\) non-overlapping tuples as,

\begin{equation}
    X = [x_1,x_2,...,x_T],x_i \in \mathbb{R}^ {C_0\times n\times V_0},
\end{equation}

where n represents the number of frames in a tuple. Then it is reshaped as follows.
\begin{equation}
    X_0 \in \mathbb{R}^ {C_0\times T\times n\times V_0} \Rightarrow \mathbb{R}^ {C_0\times T\times V},
\end{equation}

such that, the long input sequence is divided into tuples and the model is forced to learn inter-relationships within tuples and relationships across tuples, where \(T\) = \(T_0 / n\), \(V = n\times v_0\).

Finally, this sequence is fed to a feature mapping layer consisting of one convolution layer with BatchNorm and ReLU function. We avoid linear layers in the feature mapping layer to maintain the efficiency of the encoder regardless of the input sequence length.

\begin{equation}
    X_0 \in \mathbb{R}^ {C_0\times T\times V} \Rightarrow \mathbb{R}^ {C\times T\times V},
\end{equation}

Where \(C\) is the number of output channels.

\noindent\textbf{Positional Encoding:} Since transformers do not contain recurrence or convolution, there should be some mechanism to maintain the order of the joint positions such that the identity of the joints across the frames can be distinguished. Similar to \cite{attention}, we used the positional encoding module which utilise sine and cosine functions with different frequencies to encode the inputs.

Formally, let \(pos\) be the position of the joint and \(i\) be the dimension of the position encoding vector we generate the positional encoding $PE$ for each joint in a tuple such that, each joint is assigned a unique ID. The positional encoding has the same dimension \(c_{in}\)
as the embedding.
\begin{equation}
    \begin{split}
        PE(pos,2i) = \sin{(pos/10000^{2_i/c_{in}})},\\
        PE(pos,2i+1) = \cos{(pos/10000^{2_i/c_{in}})}
    \end{split}
\end{equation}

\subsection{Physics-based decoder}\label{sec:physics_decoder}

The physics-based decoder utilise laws of physics to generate the future skeleton sequences and is mainly formed with two parts, a physics solver followed by a transformer decoder. Specifically, we aim to utilise the physics-based decoder to model the TD behaviour where it generates the future skeleton sequences by utilising the encoded joint positions, forces and physics-based theories (which is encoded in a physics solver). As such, the generated predictions follow the behaviour of TD subjects.

In classical mechanics, a set of parameters could be defined to fully describe the state of a physical system in terms of positional coordinates at any given time. The equations of motion for this system are derived in terms of generalized coordinates, and they specify how the coordinates vary with time. The proposed physics-based decoder, $D$, leverages this formulation to predict future motion. It is motivated by the fact that the normal behaviour of the subject can be described by the generalised coordinates. However, the physics-based decoder will not account for the abnormal behaviour of the subject.  

The differentiable physics solver, $E$, follows Lagrangian dynamics so that it guarantees the output representations of the decoder are plausible physical positions and forces. The functionality of the physics solver can be written as,

\begin{equation}
    E(q_t,\dot{{q}_t},\textit{f},\mu) = [q'_{t+1},\dot{q}_{t+1}],
\end{equation}

where \(q_t\),\(\dot{{q}_t}\),\(\textit{f}\),\(\mu\) are the current position, velocity, control forces, and inertial properties, respectively. The inertial properties \((\mu)\), and velocity \((\dot{{q}_t})\) are gained by fitting joint data to a general skeleton model.

Then the predicted next position, \(q'_{t+1}\), is generated by integration using,

\begin{equation}
    q'_{t+1} = q_t + \Delta{t}{\dot{{q}_t}},
\end{equation}

where the scrutinized time interval is denoted by \(\Delta{t}\). 

The Lagrangian dynamic equation is solved in the generalized coordinates by the decoder to solve \(\dot{q}_{t+1}\).

\begin{equation}
M(q_t,\mu)\dot{q}_{t+1} = M(q_t,\mu)\dot{q}_{t} - \Delta{t}(c(q_t,\dot{{q}_t},\mu) - \textit{f }) + J^T(q_t)\tau,
\end{equation}

where \(M\), \(c\), \(\tau\) are the mass matrix, Coriolis and gravitational force, contact force in the generalized coordinate system respectively, with Jacobian matrix \(J\). \(\tau\) is gained by solving the linear complementarity problem (LCP) \cite{differentiable_physics}:

\[find \quad {\tau},\textit{v}_{t+1} \quad  such\quad that \]
\begin{equation}
 \quad {\tau}>0,\textit{v}_{t+1}>0,\tau^T\textit{v}_{t+1} = 0
\end{equation}

The velocity \(\textit{v}_{t+1}\) can be written as a linear function of \(\tau\):

\begin{equation}
\textit{v}_{t+1} = J\dot{q}_{t+1} = JM^{-1}(M\dot{{q}_t} - \Delta{t}(c-\textit{f}) + J^T\tau) = A\tau + b, 
\end{equation}

where \(A = JM^{-1}J^T \) and \(b = J(\dot{{q}_t} + \Delta{t}M^{-1}(\textit{f - c})) \).

Then \(A,b\) are mapped to the contact force \textit{f}.

\begin{equation}
\textit{f}_{LCP}(A(q_t,\mu),b(q_t,\dot{{q}_t},\textit{f},\mu)) = \tau 
\end{equation}

Then the output from the physics solver, the generalized positions of joints of the sequence is converted again to Cartesian coordinates and re-scaled to the input scale by the proposed transformer decoder network, $f_{phy}$, where

\begin{equation}
    {\hat{q_{t+1}}} = f_{phy}(q'_{t+1}),
\end{equation}
and $f_{phy}$ is the physics-based decoder.

These predictions of the decoder are bound to be physically realistic future poses as we utilise a differentiable physics solver within it. 

\subsection{Non-Physics-based Decoder}\label{sec:non_physics_decoder}

The non-physics-based decoder is trained to generate the future skeleton sequence of the subject by only considering the encoder's positional output. Therefore, this branch does not consider the physical feasibility of the predicted future poses. The predictions are purely driven by the learned mapping between the current position, $q_t$, and the predicted future position, ${\check{q_{t+1}}}$, where the non-physics-based decoder network try to minimise the error between the predicted future poses of the subject and his or her actual poses. This could be written as, 

\begin{equation}
    {\check{q_{t+1}}} = f_{non-phy}(q_t),
\end{equation}
where $f_{non-phy}$ is the on-physics-based decoder.

\subsection{ADOS-classifier}\label{sec:classifier}

Using the encoded generalized position and force sequences, the classifier aims to predict the ADOS score for each sequence. A feed-forward network with softmax output is adopted as the classifier. Specifically, the joint positions \(q_t \in \mathbb{R}^D \) and forces \(f_t \in \mathbb{R}^D \) at each frame are concatenated to form the feature vector \(W_t \in \mathbb{R}^{2D} \) and is fed to the classifier. Formally, let the classifier be denoted as $f_{cls}$, then the ADOS prediction can be generated as,

\begin{equation}
    y_{ADOS} = f_{cls}(W_t).
\end{equation}

\subsection{Loss Functions}\label{sec:loss}

We utilise three losses to govern the training of the proposed framework.

For training the physics branch, we leverage the mean-square-error between the predicted future positions of ASD subjects and the actual future positions of TD subjects. Non-physics branch is trained using the mean-square-error loss between the predicted future positions of that branch and the actual future positions of the subject. These losses could be written as,

\begin{equation}
    L_{phys} = \sum_{t}MSE(\hat{q_{t+1}},q_{t+1}),
\end{equation}
and
\begin{equation}
    L_{Nonphys} = \sum_{t}MSE({\check{q_{t+1}}},q_{t+1}),
\end{equation}

where $q_{t+1}$ is the actual ground truth position of the subject in frame $t+1$.

A third loss is used to discriminate between the predictions from the physics-based branch and the non-physics-based branch. Specifically,

\begin{equation}
    L_{phys-NonPhys} = \sum_{t}MSE(\hat{q_{t+1}},{\check{q_{t+1}}}),
\end{equation}
calculates the total discrepancy between the physics-informed predictions and non-physics-based predictions. The motivation is to maximise this discrepancy such that the network will be able to better understand natural human behaviour and autism-related behaviour. Then the total loss is calculated by,

\begin{equation}
    L_{total} = L_{ADOS} + L_{phys} + L_{Nonphys} + L_{phys-NonPhys},
\end{equation}

where $ L_{ADOS}$ is the binary classification loss of the ADOS classifier. 

\section{Experiments}\label{sec:eval}

\subsection{Datasets}

Multimodal Dataset for Autism Intervention Analysis (MMASD) \cite{mmasd} is a dataset that includes 1315 video clips of 32 different children (27 males and 5 females) aged between 5 and 12 years, diagnosed with autism of different severity levels. It comprises more than 108 hours of video in total. The data set has been categorized into eleven activity classes depending on the conducted activity. The demographic information and the Auism Evaluation Scores (ADOS) of all participating children have been reported along with the date of birth, motor functioning score, and severity of autism. According to the documentation of the MMASD dataset the ADOS comparison scores rage from 5 to 10. Score \( < \) 5 falls to the TD category, 5-7 as moderate ASD level and 8-10 as severe ASD level. In this paper, we use the 2D skeleton sequence with the ADOS comparison scores ranging from 6-10 in the MMASD.

Development of Robot-Enhanced therapy for children with AutisM spectrum disorders (DREAM) \cite{dream} is another dataset constituting behavioral data recorded from 61 children aged between 3 to 6 years, diagnosed with ASD. The developers have collected the samples of the subjects whose ADOS scores range from 7-20, performing three different tasks: imitation, joint attention, and turn-taking in a therapy environment where they interact with either a human therapist (SHT) or a robot (RET). With a median length of 32 minutes, each session ranged in length from 3 to 87 minutes.

To further illustrate the utility of the proposed framework in additional application domains we have conducted a third evaluation using the UR Fall Detection Dataset (URFD) dataset \cite{urfd}  for the fall prediction task. Details of this dataset and the results are provided in supplementary materials. 

\subsection{Implementation Details}

\begin{table}[tbp]
\centering
\resizebox{\linewidth}{!}{
\begin{tabular}{|l|c|c|c|c|c|c|}
\hline
Dataset & Input Length  &  Predicted Length & Joints  & Batch Size & Learning Rate  &   Epochs  \\ \hline
MMASD   & 64 & 16 & 25     & 8     & 0.001  & 50  \\ \hline
DREAM   & 64 & 16 & 10 & 16     & 0.001    & 250     \\ \hline
URFD    & 32 & 32 & 17  & 8     & 0.0001    &   10  \\ \hline
\end{tabular}}
\caption{Settings for each dataset}
\label{table:split}
\end{table}

We used the 2D-Openpose data provided in the MMASD dataset \cite{mmasd} and the raw 3D skeleton data provided in the DREAM dataset \cite{dream} as the input to the proposed model for the respective datasets. Each skeleton sequence is meanly normalised and the frames without skeletons are replaced with the skeleton from the immediately preceding frame. Then all the frames within a data sample were re-scaled with respect to the first frame of that data sample. The scale of ADOS levels differs with the dataset, however, we utilise the ground truth ADOS score levels provided by the authors of the dataset and do not change these levels.

The MMASD dataset consists of three skeletons per frame, namely the skeleton of the child, the skeleton of the therapist, and the skeleton of the interaction partner. Out of the three skeletons, the skeleton sequences of the child and the interaction partner were leveraged in our work. Therefore, the input for the proposed model is the observed portion of the skeleton sequence of the child while the target skeleton sequences for the physics branch and non-physics branch are constructed using the unseen future skeleton sequences of the interaction partner and the child, respectively.

As the DREAM dataset contains only the skeleton sequence of the child, the target skeleton sequence for both physics and non-physics branches are the unseen future skeleton sequences of the child. As such, only for the DREAM dataset, the discriminative loss between physics and non-physics branches is not utilised.

The proposed framework was implemented using the PyTorch library \cite{pytorch}. We use Adam optimizer \cite{adam} for training the models. The input size depends on the number of input frames, and the number of joints within a single frame. These details are provided in Table \ref{table:split}. Following \cite{zahan2023human} for all experiments we use 10 fold cross-validation evaluation protocol. Hyperparameters learning rate, and batch size are experimentally evaluated using the validation split of the dataset which is constructed by selecting 10\% of the data from the training split as the validation set.

\subsection{Evaluation Metrics}

As evaluation metrics, we use precision, recall, accuracy, and F1-Score. Prior works \cite{zahan2023human, VideoGestureAnalysisForASDDetection } have only used accuracy as the performance metric. While accuracy indicates how often the model is correct on average, this measure could be biased toward the majority class. In contrast, the precision metric indicates how well the model could avoid False Positives (FP) and detect True Positives (TP). In addition, recall indicates how many true positive cases the model has been able to detect. Avoiding False Negative (FN) and FP classifications is extremely important in biomedical applications, as such precision, recall, and F1-score are used as additional evaluation metrics. These metrics can be calculated as follows:

\begin{equation}
Precision = \frac{TP}{TP + FP},
\end{equation}

\begin{equation}
Recall = \frac{TP}{TP + FN},
\end{equation}

\begin{equation}
Accuracy = \frac{TP + TN}{TP + TN + FP + FN},
\end{equation}

\begin{equation}
F1-Score = 2 \frac{Precision . Recall}{ Precision + Recall}.
\end{equation}

\subsection{Results}
In this section, we provide evaluation results of the proposed models together with the state-of-the-art methods in the MMASD and DREAM datasets. The evaluations of our model for the fall prediction task using the UR Fall Detection (URFD) dataset \cite{urfd} are provided in supplementary materials. 

\begin{figure}
    \centering
    \includegraphics[width=.8\linewidth]{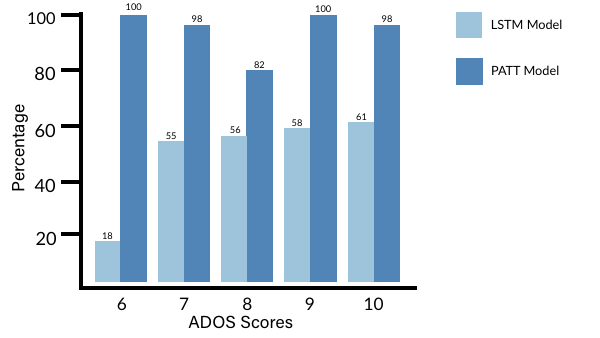}
    \caption{Comparison of classification accuracy of baseline LSTM model and the proposed PATT model for different ADOS levels on the MMASD \cite{mmasd} dataset.}
    \label{fig:mmasd_ados_bar}
\end{figure}


\begin{table}[htbp]
\centering
\begin{tabular}{|l|c|c|c|c|}
\hline
Method                            & Precision             & Recall                & Accuracy & F1 Score              \\ \hline
MSTCN \cite{wei2023vision}                        & 0.32                       &    0.29                       &    41.77     & 0.30                       \\ \hline
LSTM \cite{wei2023vision}                             & 0.54                      & 0.52                      & 58.54    & 0.52                      \\
\hline
STTFormer \cite{sttformer}                         & 0.80                      & 0.82                      & 81.64    & 0.80                      \\ \hline
PATT (ours)       & 0.96          & 0.95     & 95.89    & 0.96         \\ \hline
\end{tabular}
 \caption{Experimental Results on MMASD dataset}
\label{table:mmasd_results}
\end{table}

The quantitative comparisons between the proposed model and existing state-of-the-art methods on the MMASD dataset are provided in Tab. \ref{table:mmasd_results}. Due to the unavailability of existing baseline models that have been evaluated using the MMASD dataset, we referred to the literature on behaviour-based autism disorder detection and evaluated these methods on the MMASD dataset for baseline comparisons. Recently, in 
\cite{wei2023vision} the authors propose the use of LSTM and MSTCN models for detecting the behavioural traits in videos that are indicative of autism disorder. We adapted these models to skeleton data by changing the input shapes of those models. In addition, to provide comparisons to the off-the-shelf transformer models that have been proposed for processing skeleton data we use STTFormer \cite{sttformer} baseline. 

When analysing the results in Tab. \ref{table:mmasd_results} we observe that the proposed method has been able to achieve significant performance gain compared to the baselines. Specifically, we observe more than 35\% increase in accuracy compared to the LSTM-based model proposed in \cite{wei2023vision} and approximately 14\% gain in accuracy compared to the existing state-of-the-art transformer for modelling skeleton sequences. Figure \ref{fig:mmasd_ados_bar} compares the classification accuracy per ADOS score level on the MMASD dataset achieved by the proposed PATT model with the baseline LSTM model of \cite{wei2023vision}.  We would like to note the superior recognition of the proposed model in all the ADOS score levels compared to the baseline. 

\begin{table}[htbp]
\centering
\begin{tabular}{|l|c|c|c|c|}
\hline
Method                            & Precision             & Recall                & Accuracy & F1 Score              \\ \hline
LSTM \cite{wei2023vision}                             & 0.65                      & 0.51                       &    54.41     & 0.54                       \\ \hline
MSTCN  \cite{wei2023vision}                       & 0.40                       &    0.30                       &    30.29     & 0.31                       \\ \hline
STTFormer \cite{sttformer}                         &    0.94                       &    0.90                       &    90.68    &  0.92                       \\ \hline
GGait Model \cite{zahan2023human}                         &--                       &--                       &78.60    &--                       \\ \hline
PATT (ours)       & 0.97          & 0.93      & 95.54     & 0.95        \\ \hline
\end{tabular}
 \caption{Experimental Results on DREAM dataset}
\label{table:dream_results}
\end{table}

\begin{figure}
    \centering
    \includegraphics[width=.8\linewidth]{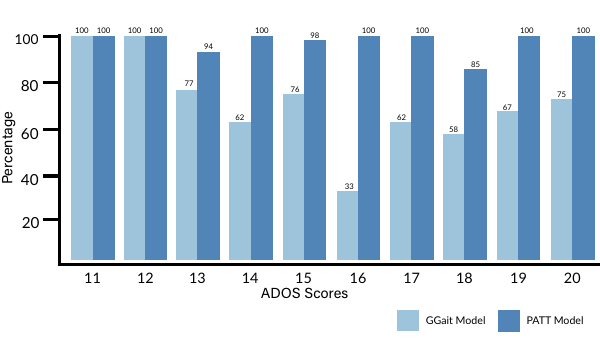}
    \caption{ Comparison of classification accuracy of baseline GGait model and the proposed PATT model for different ADOS levels on the DREAM \cite{dream} dataset.}
    \label{fig:dream_ados_bar}
\end{figure}

For comparisons on the DREAM dataset we use the LSTM and MSTCN models of \cite{wei2023vision}, STTFormer \cite{sttformer} model and the model recently proposed in \cite{zahan2023human} for autism detection. These evaluations are provided in Tab. \ref{table:dream_results}. When analysing the results we could clearly see that the proposed method has been able to achieve competitive results compared to all the baselines. Figure \ref{fig:dream_ados_bar} compares the classification accuracy per ADOS scores on the DREAM dataset achieved by the proposed PATT model together with the baseline GGait model. These evaluations clearly illustrate the superiority of the proposed model across all the ADOS score levels, despite the imbalances of the classes in the training data.   

We would like to compare our model with the state-of-the-art GGait model \cite{zahan2023human} which is a graph-convolution and transformer hybrid model. We achieve superior performance across all the ADOS score levels. Moreover, we would like to point out that the graph convolution-based MSG3D \cite{liu2020Disentangling} decoder utilised in \cite{zahan2023human} at the test time has 32M parameters which makes it parameter-heavy compared to the 100K parameters of the proposed framework at the testing setting. Therefore, at the test time, the GGait model of \cite{zahan2023human} is approximately 300 times larger than the proposed method which makes it extremely computationally exhaustive and time-consuming, which is not ideal in a clinical diagnostic setting.

\section{Ablation Experiments}\label{sec:ablation}
To gain insights into the relative contributions of the components of the proposed innovations including the novel encoder-decoder framework for ADOS score regression, the physics-based decoder, and the innovative loss function that helps discriminate between the predictions of the two decoders, we conduct a series of ablation experiments using the following models.

\begin{enumerate}
    \item \textbf{PATT-Enc: } In this ablation variant we remove the two decoder branches from our PATT framework. The encoded skeleton sequence is directly classified using the ADOS classifier. 
    
    \item \textbf{PATT-Phy Dec: } We add a physics-based decoder to the PATT-Enc model. 
    \item \textbf{PATT-Non-Phy Dec: } We add a non-physics-based decoder to the PATT-Enc model. 
    \item \textbf{PATT-w.o Dis Loss: } This is the proposed PATT model without the proposed discriminative loss which discriminates the predictions between the physics and non-physics branches. 
\end{enumerate}

Ablation experiments are conducted using the MMASD dataset and the results are provided in Tab. \ref{table:ablation}. 

\begin{table}[htbp]
\centering
\begin{tabular}{|l|c|c|c|c|}
\hline
Method            & Precision & Recall & Accuracy & F1 Score \\ \hline
PATT-Enc          & 0.88          & 0.83     & 85.44    & 0.85         \\ \hline
PATT-Phy Dec      & 0.89          & 0.87      & 88.29    & 0.88         \\ \hline
PATT-Non-Phy Dec      & 0.93          & 0.89      & 90.50   &  0.88        \\ \hline
PATT-w.o Dis Loss & 0.88          & 0.88       & 87.34         & 0.88         \\ \hline
PATT (ours)       & 0.96          & 0.95     & 95.89    & 0.96         \\ \hline
\end{tabular}
\caption{Results of different ablation variants of the proposed method on MMASD dataset.}
\label{table:ablation}
\end{table}
When comparing the results of \textbf{PATT-Enc} with the proposed \textbf{PATT} model's results we can clearly see the impact of the proposed encoder-decoder architecture. We believe a significant 10.45\% gain in accuracy when the two decoder branches are added to the model. We believe this is because of the ability that the decoding branches provide to the overall framework to compare and contrast behaviour of children with and without autism disorder and identify distinguishable features which could support the classification. 

In addition, using the \textbf{PATT-Phy Dec} ablation variant we demonstrate the utility of the proposed physics-based decoder branch. We would like to compare the performance gain that the PATT-Phy Dec model achieves compared to the PATT-Enc model with the simple addition of the physics-driven decoder. Similarly, a notable performance gain is achieved with the addition of a non-physics-based decoder (i.e. \textbf{PATT-Non-Phy Dec} branch). With these decoder branches, we force the encoded latent embedding to carry information regarding the subject's behaviour, which helps the identification of ADOS levels.   

Moreover, with the addition of the proposed discriminative loss which enables the two branches to compare and contrast their predictions we further improve our performance gain. This is clearly indicated by the performance drop we observe in the \textbf{PATT-w.o Dis Loss} when the discriminating loss is removed from our framework. This validates our hypothesis that there exist discriminative behaviour patterns between children with and without autism disorder and explicitly comparing them helps further augment the learned features of the encoder. 

\begin{figure}
    \centering
    \includegraphics[width=\linewidth]{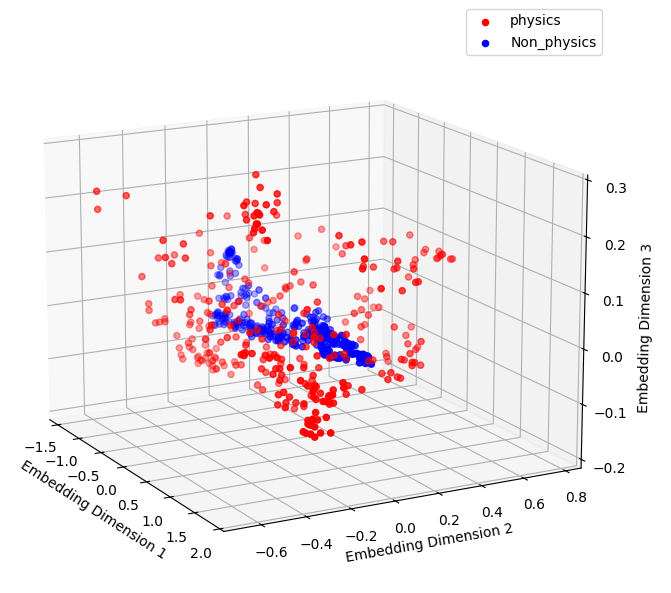}
    \caption{3D embedding space visualisation of the physics and non-physics-based branches of the proposed framework.}
    \label{fig:pca}
\end{figure}

To further illustrate the discriminative power of the proposed framework we visualise the 3-dimensional representation of the embedding spaces extracted from the 2nd last layer of the physics and non-physics decoders of the proposed model for the MMASD \cite{mmasd} dataset. This visualisation is presented in Fig. \ref{fig:pca}. To generate a 3D illustration of the higher dimensional latent space we used Principal Component Analysis (PCA).


When analysing the visualisation in Fig. \ref{fig:pca} we observe that the physics branch has more extensive embedding space, which is expected as it captures the distinct motion patterns of TD subjects. In contrast, a more compact latent space is observed in the non-physics-based branch, which is directly optimised to match the behaviour of the subjects with an autism spectrum disorder. Therefore, these results clearly validate the discriminative learning capability of the proposed framework.

\section{Conclusion, Limitations, and Future Work}\label{sec:conclusion}
\textbf{Conclusion.} In this paper, we have introduced a physics-informed novel deep learning framework for Autism Spectrum Disorder (ASD) severity recognition. We show how a discriminative architecture can be developed by incorporating laws of physics into the neural network design such that robust features are learned by explicitly differentiating between normal and abnormal behaviour patterns, allowing effective localisation of behavioral traits that are indicative of ASD. We have conducted extensive experiments using multiple publicly available benchmarks, including the Multimodal Dataset for Autism Intervention Analysis (MMASD) dataset and the Development of Robot-Enhanced therapy for children with Autism spectrum disorders (DREAM) dataset. Our results clearly demonstrate that we outperform the current state-of-the-art methods by significant margins. Our evaluations demonstrate the utility of learning of the ASD diagnosis task together with the prediction of future motion which allowed our framework to differentiate between the highly probable future poses based on laws of physics and the actual poses of the subject.
Furthermore, the proposed framework is not restricted to ASD severity recognition and can be applied to any application that requires analysis of human motion patterns. To demonstrate the wider applications of our proposed approach we have conducted a third experiment using the UR Fall Detection Dataset (URFD) dataset for the fall prediction task which further reinforces the superiority of the proposed deep learning framework for analyzing human motion patterns to detect abnormal events.
\textbf{Limitations.} While our framework is flexible for both 2D and 3D input skeleton data representations, 3D representations are recommended as they carry detailed information regarding the skeleton representations. However, this could be a limitation when extending the framework to handle coarse-grained inputs such as 2D skeleton data and other spatial inputs such as RGB videos.
\textbf{Future Work.} In our work, we will investigate how additional physics modeling can be leveraged to minimise the reliance of the framework on 3D input representations. Furthermore, we will investigate the impact of other feasible physical modeling approaches such as "Hamiltonian mechanics".

\section*{Acknowledgment}

The research presented in this paper was supported partly by an Australian Research Council (ARC) Discovery grant DP200101942 

\bibliographystyle{IEEEtran}
\bibliography{egdb}

\newpage

\section*{Supplementary Materials}

\subsection{Evaluations on UR Fall Detection Dataset for Fall Prediction Task}

In this section, we describe the performance of the proposed Physics Augmented Tuple Transformer for the fall prediction task which we evaluate using the UR Fall Detection (URFD) \cite{urfd} dataset. 

The UR Fall Detection (URFD) \cite{urfd} dataset is created by capturing videos from two different camera angles of daily living activities. It contains 70 videos in total and 30 of them are falling motions. Most of the recent literature, including \cite{noorfall, wang2024enhancing} utilise this dataset for the fall detection task. In contrast, we demonstrate our model for the fall prediction task which is more beneficial in biomedical and safety monitoring applications such as in monitoring elders in aged care facilities. Specifically, using the video sequences in the URFD dataset we generate the fall prediction task as follows. The videos in this dataset have variable lengths. From each sequence, we observe the first 32 frames and pass only those frames to our model. The two decoders try to predict the skeleton motion in the rest of the frames of the video. 

Tab. \ref{table:fall_results} provides the evaluation results together with baselines.

\begin{table}[htbp]
\centering
\begin{tabular}{|l|c|c|c|c|}
\hline
Method                            & Precision             & Recall                & Accuracy & F1 Score              \\ \hline
LSTM \cite{wei2023vision}                             & 0.70                       &    0.74                       &    78.57     & 0.71                       \\ \hline
MSTCN  \cite{wei2023vision}                       & 0.50                       &    0.14                       &    28.57     & 0.22                       \\ \hline
STTFormer \cite{sttformer}                         &    0.75                       &    0.91                       &    85.71    &  0.79                       \\ \hline
PATT (ours)       & 1.0         & 1.0     & 100    & 1.0       \\ \hline
\end{tabular}
 \caption{Experimental Results on UR Fall Detection (URFD) \cite{urfd} dataset}
\label{table:fall_results}
\end{table}

When analysing the results in Tab. \ref{table:fall_results} it is clear that the proposed method has been able to achieve superior results compared to the baselines. This clearly validates the need for the two generative decoders in the proposed architecture which allows the model to compare the poses that are needed for stable motion and poses that lead to instability. We believe this ability to discriminate the stable motion patterns together with the actual poses of the subject allows our model to identify the early cues of instability in the subjects and make accurate predictions only using a few frames. 

\end{document}